# DATA SCIENCE APPROACH TO PREDICT THE WINNING FANTASY CRICKET TEAM—DREAM 11 FANTASY SPORTS

Sachin Kumar S [1], Prithvi H V [2], C. Nandini [3]

[1] [2] Final Year Students, Department of CSE, DSATM, Bengaluru, Karnataka, India

[3] Head of the CSE Department, DSATM, Bengaluru, Karnataka, India

sachinsks1999@gmail.com [1]   hvprithvi09@gmail.com [2]

hodcse@dsatm.edu.in [3]

*Abstract*— The evolution of digital technology and the increasing popularity of sports inspired the innovators to take the experience of users with a proclivity towards sports to a whole new different level, by introducing Fantasy Sports Platforms (FSPs). The application of Data Science and Analytics is Ubiquitous in the Modern World. Data Science and Analytics opens doors to gain a deeper understanding and helps in the decision making process. We firmly believed that we could adopt Data Science to predict the winning fantasy cricket team on the FSP, Dream-11. We built a predictive model that predicts the performance of players in a prospective game. We used a combination of Greedy and Knapsack Algorithms to prescribe the combination of 11 players to create a fantasy cricket team that has the most significant statistical odds of finishing as the strongest team—thereby giving us a higher chance of winning the pot of bets on the Dream-11 FSP. We used PyCaret Python Library to help us understand and adopt the best Regressor Algorithm for our problem statement to make precise predictions. Further, we used Plotly Python Library to give us visual insights on the team, and player's performances by accounting for the statistical, and subjective factors of a prospective game. The interactive plots help us to bolster the recommendations of our predictive model. You either win big, win small or lose your bet based on the performance of the player's selected for your fantasy team in the prospective game, and our model increases the probability of you winning big.

## I. INTRODUCTION

The term "Sports Analytics" gained widespread popularity in the field of Sports as well as in the field of Statistical Analysis following the release of a book titled "Moneyball: The Art of Winning an Unfair Game," by Michael Lewis [1], in the year 2003 and this popularity surmounted when the movie "Moneyball [2]" was released in the year 2011. In the story of "Moneyball," the Oakland Athletics General Manager, Billy Beane, who was also a former ballplayer, uses analytical methods to derive insights from the Player's historical data and makes pivotal decisions for his team, the Oakland A's. He relied intensely on the application of analytics to construct an aggressive baseball team on a minimal budget, and his team set the world record for winning 20 consecutive games in the history of Baseball. After the success of Oakland Athletics in the game of Baseball by using Sabermetrics, an analytical approach adopted by Billy Beane to select low-cost players with high on-base percentage, many Baseball teams jumped-in on the strategy. This breakthrough success of Analytics to make decisions in Sports made people adopt it not only in Baseball but also in Basketball, American Football, Soccer, Tennis, Cricket, and every other major, highly popular organized sport.

Sports Analytics came into existence with the experimentation of Statistical Analysis methods on the data of Baseball in the 1970s by the statisticians of the Society for American Baseball Research (SABR). After its proven success by Billy Beane in the field of Baseball in optimizing player's and team's performance, people worldwide started experimenting with the datasets of Football, Hockey, Basketball, Soccer, Tennis, and Cricket. In the modern world, Sports Analytics is found to be used in almost every organized sport that is played. Today, we have Sports Analytics put into use in all primary sports right from Team-Selection and On-field decision making to business aspects of the sport. The development of this domain had its roots primarily from Statistics, Game Theory, and Decision Sciences, and today, the field also uses Machine Learning and Modern Analytical Approaches to drive decisions for the team and the game itself.

With the application of Sports Analytics, we can get insights on the Sports Market, Assess Players and their Performances, rank participating teams, Predict Player's score and team's score in a prospective game, Make Game Day Decisions to deliver a win, Promote Team Brands and improve the team's income, optimize and manage team's budget and finances. With such a wide range of applications, we have to consider the varied methodologies for each of these applications and select the best methods to maximize the outcome of the application of analytics. Literally, in the modern world, every decision in every major sport is made based on analytics, but this knowledge is highly esoteric and not known to most of the people who are not into the field of Analytics related to Sports and Sporting Events.

Sports Analytics is the process that involves gathering all relevant historical statistical data of players to develop a hypothesis based on statistical analysis and generate insights that come handy in making decisions for the team. There are two primary aspects when it comes to Sports Analytics, and these aspects apply to all kinds of Organized Games, and they are: "on-field analytics" and "off-field analytics." On-field analytics deals with the improvisation of the Player's performance or the team's performance on the field using insights from analytical reports. Off-field analytics deals with the business and market of sports. This becomes useful when the team has to consider a particular budget for a tour and pick players who suit the team's total budget for the tour. Sports Analytics is a more generalized term, and its application, although similar, has different approaches and outcomes in various sports. Usually, in every game, there is a blend of both the aspects in making decisions for a Team Selection and also for another decision making related to the team. In our case, to predict a winning fantasy cricket team, we employ both On-field and Off-field analytics for decision making. On-field helps us identify the best performers in the given set of "n" players, Off-field analytics help us pick 11 top players suiting our budget of 100 credit scores as set by Dream-11 FSP.

In the field of Sports Analytics concerning any sport, the most quintessential requirement to generate better analytics with the available empirical data on the competition is by asking the right questions and using the correct methodology to answer those questions. There can be more than one way to solve a problem, but the most desirable and efficient way always yields better results and better predictions. In cricket, there is a plethora of data available on the players and their strengths and weaknesses against rival teams and their performances in various pitches around the world. To make a Team selection or decision for the team, asking the right questions is a start, and answering those questions with the right tools and most-proven methods is a necessity because the most reliable tools and techniques yield better predictions.

Building winning sports teams and successful sports businesses are more likely when data and models guide the decisions. Sports analytics is a source of competitive advantage. To apply the analytics feasibly to gain a competitive advantage in sports, we should initially grasp the sport itself—the sports industry, the organized sports business, on-field factors and incidents, and courts of play. We have to envision and discern how to work with the data if sports recognize viable and verified data sources, gather relevant datasets to extract useful information, arrange, and model them to make it viable for investigation. Also, we need to learn from the data to consider how to assemble models from the information. Data cannot represent themselves. Informative pivotal forecasts do not emerge out of thin air. We must gain from information and build models that work.

## II. LITERATURE REVIEW

There is a plethora of literature available on Sports Analytics and its application to make decisions for sports like Baseball, Football, Basketball, and Soccer, and Tennis. However, there exists only a fewer number of pivotal research articles on the use of Sports Analytics in the Sport of Cricket. In this section, we will give you an overview of the evolution of Sports Analytics from Baseball to Cricket, and different methods employed by different authors to make outcome predictions and team selections in the field of sports.

To introduce anyone to the vast field of Organized Sports Analytics, the best place to start with is the book "Moneyball: The Art of Winning an Unfair Game [1]", a book by Michael Lewis, published in 2003. The story of this book revolves around the General Manager of the Team, Oakland Athletics, Mr. Billy Beane, who set a world record by delivering 20 consecutive wins to the team with a minimal budget. The team was suffering from the shortage of funds, and Mr. Billy Beane adopted the principle of Sabermetrics, an Analytical approach to choose players for the team who cost relatively way less in price and had a more significant on-base percentage and slugging percentage. His choice of the on-base percentage and slugging percentage as better indicators of offensive success was by the use of rigorous statistical analysis. The Oakland A's' victory in the MLB Tournament in 2002 is a history that will last as long as the field of Sports and Sports Analytics holds water. James came up with the invented model the Sabermetrics, the year 1980 to sketch the science of analytics applied to the sport of baseball in honor of the Society for American Baseball Research (SABR), which was established in the year 1971. Yet, this analytical tool was not applied in practice because the Organization and members of the baseball team selection believed that a statistical tool could not surpass their years of experience. However, in the 2002 MLB Tournament, they were disproved by Billy Beane's use of Sabermetrics for selecting his Oakland A's Team. Beane's success was a result of Alderson's recommendation to Beane: they recognized which statistical tool most closely correlates the scoring of runs and other factors of the sport with the winning of the game, and at the same time the same factors that were undervalued by the rest of the baseball teams to make a selection decision. His analysis was that a player with an on-base percentage of .295 was paid around 4 million USD and a player who showcased an on-base percentage of .260 was paid only around 200 thousand USD. However, in the actual game setting, this overrated measure did not make much of a difference in delivering the most anticipated victory to the team all the time. Therefore, based on this analysis and the intuition, Billy Beane hired the player with an on-base percentage of .260 for a price that is a throw-away price when compared to buying a player with an on-base percentage of .295. Following the success of the Oakland A's in the MLB Tournament, analytics departments emerged in all the baseball team front offices.

Although the advent of Sports Analytics began with the game of Baseball in the early 1980s, the earliest organized sport was the Game of Cricket. We have reliable data on the scorecards of the games played since 1697 in the game of cricket. The record-keeping of games is what that gave rise to the introduction of Statistics and Analytics into the field of Sports. Of all the sorted out sports appreciated in America, it's maybe not astonishing that the one quintessential game to involve numbers as a prominent piece of its soul would be Baseball. Each occasion that happens in a ball game does as such with barely any undisclosed amounts: there are always nine defenders in the similar general territory, one-hitter, close to three base sprinters, close to 2 outs. There is no persistent whirlwind of action as in b-ball, soccer, or hockey; the occasions don't depend vigorously upon player arrangements as in football; there are no turnovers and no clock—each play is a discrete occasion. Thus, the advent of Sports Analytics took place with the game of Baseball. In spite of the availability of data for the game of cricket since 1697, the actual application of Analytics into organized sports happened with the game of Baseball. The most prominent reason for this unsung popularity of the sport, cricket in the field of Analytics was because it was not so popular in America. It was Americans that started applying statistics to sports data and Baseball was the most famous American Sport.

Spots that display the quintessential characteristic of placing two teams on a field of some size with a ball that is advanced towards a goal line has been around for as long as the human civilizations. The advent of the majestic game football is evolutionary and the fact that there exist different versions and variations of the game Football is perplexing. By the era of the 19th century, football was the game that was cherished by all the schools and universities across the world and prominently in the American, European and African Countries. With no authority governed by a revered central board at the time, each major school and/or university adopted a different variation of the sport depending on the surroundings of the university and the land that was available. Hence, we have the Association Football or the Soccer version of Football, and the rugby version of the football. In spite of these variations of the game, there were rudimentary data that was recorded judiciously since 19th century. Thus the application of Sports Analytics is somewhat viable in these games. In 1979, the revered organization, NCAA introduced an efficient metric called the "passer efficiency rating" as a reliable means to evaluate quarterbacks beyond touchdown passes and total yard and in the modern football games, all sorts of data are being recorded. Thus, the application of Analytics to these games are delivering more satisfactory results.

The third American sport that saw the application of Sports Analytics in recent years was Basketball. Like the game football, the game basketball was also cherished by colleges and imminent universities before the sport turned into an international organized sport with over a billion fans revering and following the sport in their routine lives. NBA Analysts are one of the highest-paid in the world and they run analytics on the game of Basketball to make decisions for the NBA Teams. In the information-rich era of the present-day modern world, the relationship between sports and numbers is closer than ever before. With the ease of access to all information publically available to anyone with internet access, it is easier than ever before to get the data about any organized sport and analyses can be run by anyone interested with a spreadsheet application to derive potential insights about the game from form of the game to the numbers that talk about the business of the sport. It is no longer as difficult as it was in the 1970s when people had to refer the newspaper to get statistical information about the sports and use a pocket calculator to perform statistical analysis on the aggregated information and report them with a manual typewriter.

The one paper that served as the base for conducting our research is Increased Prediction Accuracy in the Game of Cricket using Machine Learning [12]. In this paper, the author states that Player Selection is the most pivotal task that depicts the chances of a win or loses in the game of cricket. He considers various factors, such as the opposition team, the venue, his consistency, and current form (player statistics), etcetera, and mentions that the players' performance depends on these variables. In his paper we understand that the team management, the coach, and the captain select the best playing 11 players for each match from a squad of 15 to 30 players by analyzing the statistics and characteristics of the players. This paper [12] attempts to classify the performance of Batsman and Bowlers in One Day International (ODI) matches into five classes labeled from 1 to 5: 1 being the least performers category and 5 being the top performers. The author mentioned the use of four Multiclass Classification Supervised Learning Algorithms- Naïve Bayes, Random Forest, Multiclass SVM, Decision Trees- in this paper. The paper predicted the class to which a player belongs to in a prospective game by predicting the range in which a batsman will score runs, and a bowler might take wickets. The paper suggests that Random Forests Classifier Model provides better predictions than the other three classification models he adopted. In his research, the Random Forests showed 90% accuracy in predicting Batsman's performance and 92% in predicting bowler's performance. The author did not use the regression models to predict player's performance. Instead, he grouped the runs scored and wickets taken into five buckets with a considerable range and
predicted in which class a player might stand in the prospective game. The author used only ODI Player's dataset. This paper allowed us to question the use of classifier models and test out Regressor Models on Player's our gathered player's performance dataset of 3100 matches in three-game formats: ODI, IPL, and T20.

Some trends and articles that inspired the authors of [12] are: Muthuswamy and Lam [21] predicted the performance of Team India's current bowlers (In 2008) against the top seven international cricket teams using backpropagation network and radial basis network function. They predicted how many runs an Indian bowler might concede and how many wickets he might take in a prospective ODI match. Barr and Kantor [22] proposed criteria for comparing a pool of recommended players and selecting batsmen for the team in limited-overs cricket. The criterion suggests players who are likely to hit crazy boundaries and help the team or at least get out and offer other players a chance when they find it challenging to give their best. Iyer and Shard [23] showed how to use neural networks for predicting player performances and classify the batsmen and bowlers separately into three optimal categories – performer, moderate, and failure. This ideology served as the base for the authors of [12] to classify the players into five buckets. Surprisingly, our base paper authors neglected to use Neural Networks. However, in our case, we believed that

Neural Networks Training time is very high. So, we stuck with Regressor Algorithms without exploiting the capabilities of a Neural Network. Next, Jhanwar and Paudi [24] predicted the outcome of a cricket game by comparing and analyzing the strengths of the two participating teams. Finally, Mukharjee [25] applied Social Network Analysis to rate batsmen and bowlers in team performance. Here, he predicted the outcome of a player's performance as High, Average, and Low based on his popularity on Twitter. The author tried to find a correlation between popularity and performance, which is exciting but not practical.

The literature that gave us confidence in using subjective factors of the game for predicting player's performance in a prospective game was authored by Kalanka P. Jayalath [26]. Unpredictable variables such as Home-field Advantage, Coin-toss Result, Bat-first or second, and Day Gave vs. Day-and-Night game format have some impact on every cricket game. The author of A Machine Learning Approach to Analyze ODI Cricket Predictors [26] attempted to analyze the impact of these unpredictable variables on ODI Cricket matches by quantifying their significance. The author used three models in quantifying the significance of unpredictable variables. They are Logistic Regression Analysis, Classification Trees, and Regression Trees. The paper proved that Unpredictable variables do have a pivotal impact on the outcome of every cricket game. The quantitative value of this impact was very high on some teams, while other teams did show some that these variables mattered on a scale of moderate to less. The author did generalize the concept of Home-grounds to Continents to which a team belonged. However, in our research, we were confident to adopt the Stadium Location as a metric feature for determining players' performance.

We would also like to acknowledge the work of researcher Tinniam V Ganesh for publishing his package-yorkpy [27] and nine articles [28] to [36] on how to use effectively use the same. Yorkpy is built to be used for analyzing performances of cricketers based on match data from Cricsheet. Yorkpy helped us enormously in procuring the usable CSV data tables in our python environment by performing feature engineering, and data transformations on the 3100 cricket match YAML files downloaded from cricsheet.org [40]. Yorkpy's several other functions helped us gain a more in-depth understanding of Sports Analytics applied to Cricket. Also, we would like to mention that the author, Tinniam V Ganesh, responded to us when we reched out to him personally, and he clarified by helping us understand his work, and resolved issues we faced while using his package. It was a serendipitous encounter for us when a function of his package that was intended for conversion of T20 format cricket matches YAML files worked flawlessly when used for converting ODI game format YAML files. When we analyzed how it worked, we found that the function for coverting YAML files to CSV "convertYaml2PandasDataframeT20" was a pure function that was highly generic, which motivated us to write functions that are pure functions and abstract.

## III. METHODOLOGY

In this paper, I'd like to split the methodology section into six sub-sections: Getting Data, Feature Engineering, Data Visualization, and Machine Learning, Data Engineering, and Dream Team Recommending module.

Before I help you cognize the approach adopted to develop our product for predicting the statistically undefeatable fantasy cricket teams in ODI, IPL, and International T20 format cricket games, I'd like to give you a brief architecture of our product and explain it in seven steps.

All the sub-sections except the Machine Learning sub-section will be briefly explained in the Architecture sub-module, and you will be introduced to the Machine Learning sub-module in detail after the Architecture section.

### A. ARCHITECTURE

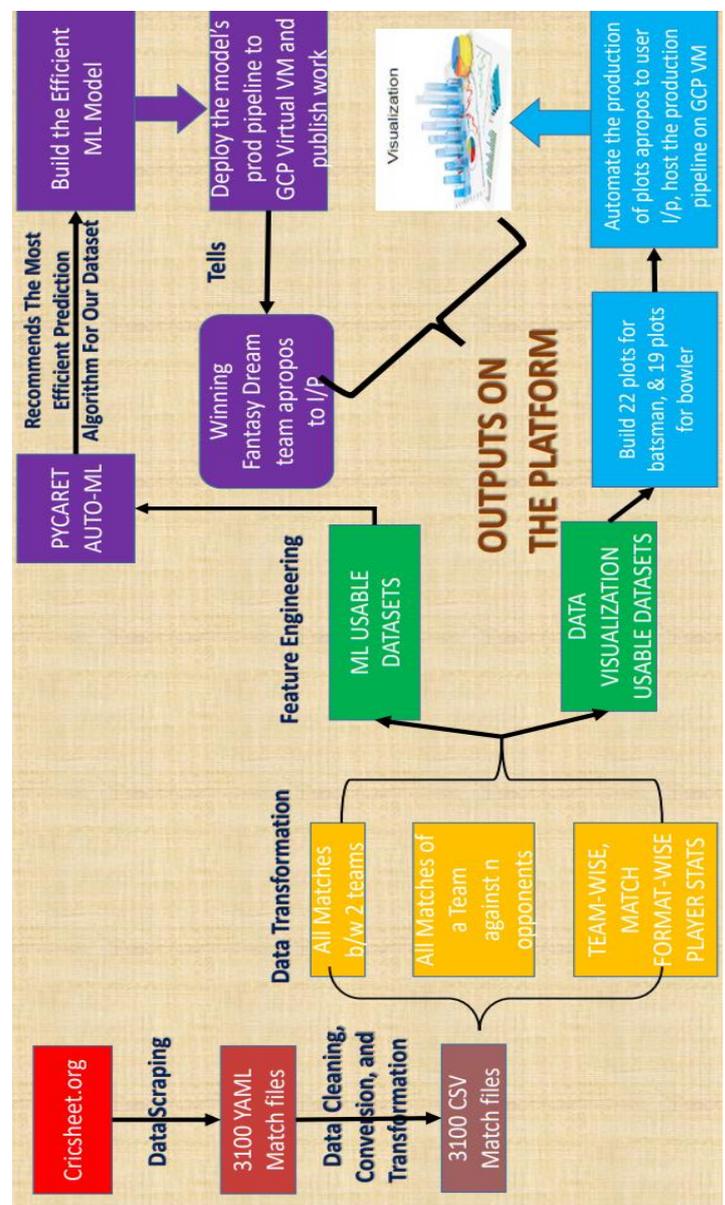

**Fig 3.a: Architecture of the Product**

## 1) *Identify Reliable Data Sources*

To solve any data science problem, data is quintessential requirement. We began by reviewing the research literature in the field, blogs & articles, and learnt what others have done in the past pertaining to the field of Fantasy Sports, Sports Analytics, and Predictions of Player performances in Cricket and Dream-11. Then we identified relevant data sources used by the authors of the literature for analysis and modeling. In our case, the base paper had used data from espncricinfo [38] website. But, the dataset had player's overall statistics, and we were looking for player's performance data in each match he played recorded as a single row. There was no ready-made dataset available for our problem statement. However, we found the reference to Cricsheet [37] in Yorkpy [27] package's github repository. We found ball-by-ball dataset for every cricket match in ODI, T20, and IPL format recorded since the year 2000. Yet, we had to perform feature engineering on this dataset to get playe's performance data in every match he had played as a single row of the team's data table.

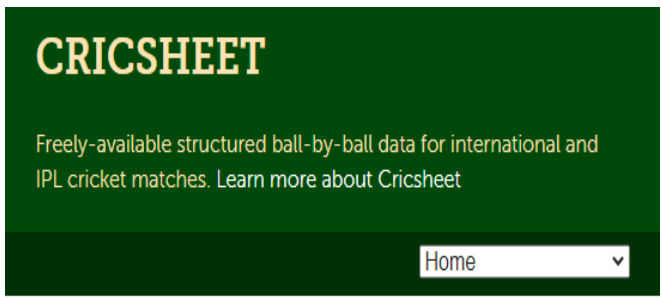

**Fig 3.a.1: Data Source (CricSheet.org)**

## 2) *Data Scraping, and Data Extraction*

We identified reliable data source at cricsheet.org where we got dataset of 3100 cricket matches files (1529 ODI, 756 IPL, 815 T20) of past two decades. The files were available in YAML format, and we scrapped it and extracted the data into CSV files using the already available specific functions to do the job in the YorkPy Python Package.

```
211028 - Notepad
File  Edit  Format  View  Help
---
meta:
  data_version: 0.9
  created: 2013-02-22
  revision: 1
info:
  city: Southampton
  dates:
    - 2005-06-13
  gender: male
  match_type: T20
  outcome:
    by:
      runs: 100
    winner: England
  overs: 20
  player_of_match:
    - KP Pietersen
  teams:
    - England
    - Australia
```

**Fig 3.a.2.1: Snippet of Scrapped YAML Data**

```
File no= 1267
Converting file: 667641.yaml
Team1:diff: ['byes', 'penalty', 'noballs']
team1: byes
team1: penalty
team1: noballs
Team2:diff: ['byes', 'penalty', 'noballs']
team2: byes
team2: penalty
team2: noballs
Team1-missing columns: []
Team2-missing columns: []
Length of info field= 11
Wintype= runs
Win value= runs
win margin= 24
../../csv_files/ODI/ODI-Matches/New Zealand-India-2014-01-19.csv
Dataframe shape= (610, 38)
```

**Fig 3.a.2.2: Snippet of Transformation logic output converting a single YAML file to CSV file**

As you can see in Fig 3.2.1, the transformed dataset has 38 columns and it is impossible to attach a snippet of the same in this document. We have effectively converted all the 3100 YAML Files to CSV in this task. All the transformed CSV files are ball-by-ball dataset of shape (no.of balls played in the game, 38).

## 3) EDA, Data Transformation

We performed Exploratory Data Analysis on a single match CSV file from each Match Format. We found out that the files contain ball-by-ball dataset of each match, and we needed the player's performances dataset in every match he played. So, we transformed all the files grouped by player, and team names, and generated the following fields:

**For Batsman: (Each batsman of the team)**
Runs, Balls, 4s, 6s, 50s, 100s, Duck Out, Strike Rate, Rival, Venue

**For Bowler: (Each bowler of the team)**
Overs, Runs Conceded, Maidens, Wickets, Economy Rate, Rival, Venue

Then we appended all the files that generated data for batsman and bowler into a single file with name of the format:
TeamName-BattingDetails.csv
TeamName-BowlingDetails.csv

We also used Yorkpy's functions to generate csv files of all the matches data between two teams, and all the matches played by a team against all the rivals. However, only the data generated using the former transformations were useful in our research.

**Fig 3.a.3.1: Snippet of Transformed Batsman Data**

**Fig 3.a.3.2: Transformed Bowler Data Snippet**

As you can see in this final output file we have merged all batsman's data that was generated apropos to team names and game format are merged into a single file with different columns that allows us to distinguish the team to which the player belongs to (Team1), and the game format in which the match was conducted (MF).

## 4) Feature Engineering

In this sub-module, we engineer the features like 50s, 100s, duck outs, Cumulative strike rate, 4 wicket, 5 wicket, Cumulative Economy, Moving average, and Dream 11 scores for the players in the datasets Batsman and bowlers. All these newly engineered features are appended to the dataset as a column and we use it for further for our Visualization and Machine Learning modules.

**Fig 3.a.4.1: Feature Engineered Batsman Dataset**

**Fig 3.a.4.1: Feature Engineered Bowler Dataset**

*5) Data Visualization Module*

This module effectively uses both the Feature Engineered Batsman and Bowler datasets to build effective interactive visualizations that help us to comprehend the details of player's strengths, their performances in different games, player's and team's strengths and weaknesses.

We used Plotly Python Library to generate 22 Interactive plots for batsman, and 20 Interactive plots for bowler. We also generated 10 interactive plots for the overall team insights.

Types of plots generated: Box Plots, Histograms, Pie Charts, Time-Series Charts, Dist Plots, Erroe Bars Plots, Violin Plots, and Linear & Non-linear Trend Lines.

Few examples of the plots will be showcased in the Results and Analysis Section.

*6) Machine Learning Module*

In this module we have attempted the following:
- Implemented the base paper classification models on our dataset and jusitified why regression models could be a better approach to predict player's performances.
- Created a heatmap to learn the best features for Input to our classification model.
- Used PyCaret Library to find the best regrossor model for our problem statement
- Built the Model on 0.07% of Dataset and tested it on 0.93% of Dataset.
- Created a pipeline that takes user input,, query all the rows alike to user's input, transforms and respapes it into a matrix, apply the Matrix as input to the model and derive predictions, return, the average pof the predictions to the user.

Each of the above five steps will be discussed in the following ML Sub-section in detail.

*7) Data Engineering*

There are two data Engineered pipelines in our product.
- Machine Learning Pipeline
- Data Visualization Pipeline

Machine learning Pipeline Architecture has 5 steps:
- Take 5 inputs from user
- Fetch all rows containing all the five inputs from both batsman and bowler datasets and recursively delete each input and fetch the rows.
- Transform the list of batsman rows into a n*13 matrix, and the list of bowlers into a n*12 matrix.
- Feed the matrices to its respective models, and derive n predictions.
- Then, take average of the n predictions and return it to the user on the UI.

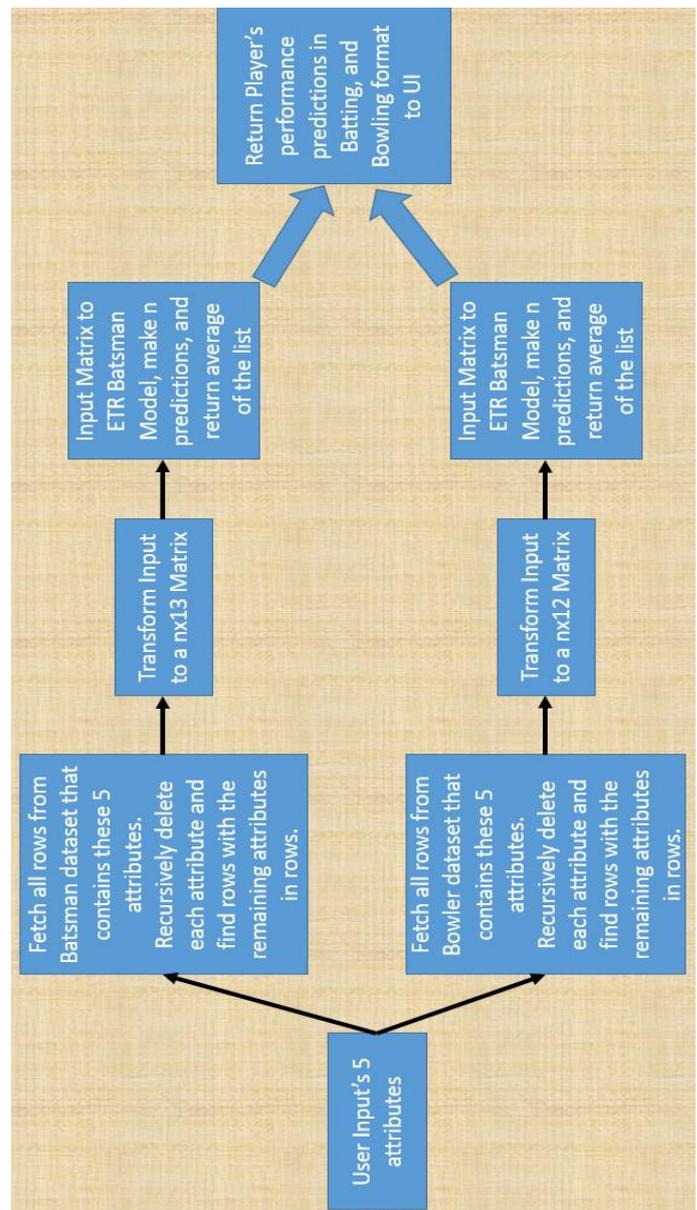

**Fig 3.a.7.1: Machine Learning Engineered Pipeline**

Data Visualization pipeline has the following steps:
- Take inputs from user on the team he wants to generate interactive visual insights.
- Feed the input of team name to the team plot function that generates 10 plots on the team
- Query the database and find all batsman and bowlers of the team, append player names to a list.
- For each player on the list generate 22 interactive plots for batsman, and 20 interactive plots for bowler.
- Display all the 52 plots in 3 windows on a tab, each tab has the ability to display one plot at a time and help user navigate to the next with buttons "previous", and "next".

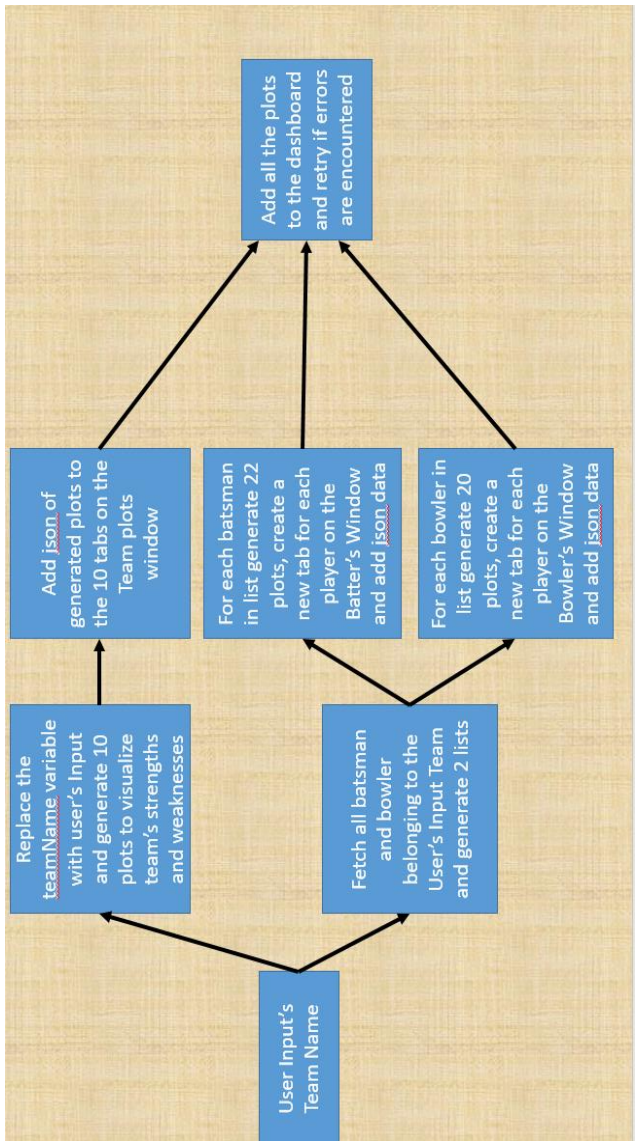

**Fig 3.a.7.2: Data Visualization Engineered Pipeline**

Further, the Team recommendation module is called as the last step in the Machine learning pipeline. This event has no trigger associated and is an independent module. Hence, it is not mentioned in the Machine learning pipeline architecture.

Tools used for creating the Pipelined architecture are: Flask, and Airflow to run the DAGs on the GCP VM.

### 8) *Fantasy Dream Team Recommendation for the standard 100 Credits Budget*

In a prospective game between two teams, we have 22 participating players. However, for the selection of our fantasy dream team, we have the following conditions:
1. Cannot select more than 7 players from one team
2. Cannot exceed the credit limit of 100 points.
3. Must select 11 players for your team

To perform this selection 11 players within the budget we used a combination of Knapsack and Greedy Algorithms. Our selection algorithm recommends you players for your team such that the expected dream-11 score of your team is maximum, and the credits used for selecting 11 such players will cost you <=100 credits.

```
def customknapSack(W, wt, val, n, L):
    K = [[{"weight": 0, "items": []}
        for x in range(W + 1)] for x in range(n + 1)]

    # Build table K[][] in bottom up manner
    for i in range(n + 1):
        for w in range(W + 1):
            if i == 0 or w == 0:
                K[i][w] = {"weight": 0, "items": []}
            elif wt[i-1] <= w and len(K[i-1][w-wt[i-1]]["items"]) < L:
                print(K[i-1][w-wt[i-1]]["items"])
                if val[i-1] + K[i-1][w-wt[i-1]]["weight"] > K[i-1][w]["weight"]:
                    bigger = list(K[i-1][w-wt[i-1]]["items"])
                    bigger.append(val[i-1])
                    K[i][w] = {"weight": K[i-1][w-wt[i-1]]
                        ["weight"] + val[i-1], "items": bigger}
                else:
                    K[i][w] = {"weight": K[i-1][w]["weight"],
                        "items": K[i-1][w]["items"]}
            else:
                K[i][w] = K[i-1][w]
    return K[n][W]

# Driver program to test above function
n = len(val)
val.sort(reverse=True)
val = [x for _, x in sorted(zip(val, val))]
wt = [x for _, x in sorted(zip(val, wt))]
print(val, wt, valRatio)
print(customknapSack(W, wt, val, n, L))
```

**Fig 3.a.8: Team Recommendation Logic**

### B. MACHINE LEARNING SUB-SECTION

In this sub-section, we will give you a brief overview on the approaches we took to test our base-paper models, select the right model for our problem statement, and implementation details.

### 1) *Testing Base Paper Classification Models*

At first, after visualizing the player's performances we realized that bucketing the players based on their performance when we had more than one data-point of the players was not an optimal approach. We firmly believed that Regressor models is the best-fit for our problem statement. However, we had to support our theory practically and hence we choose to fit our dataset to classifier models and measure the prediction accuracy.

Surprisingly, few of the models did perform well when asked to classify players dream-11 scores in buckets of 10. The results are in the figure below.

| CLASSIFIER | ACCURACY(%) |
|---|---|
| KNN | 49.75 |
| Naieve Bayes (MNB) | 33.36 |
| Naieve Bayes (GNB) | 84.72 |
| SVM | 50.34 |
| Decision Trees | 37.56 |
| Random Forests | 92.45 |

**Fig 3.b.1.1: Classifier Model's Accuracy for predicting Batsman Dream-11 Scores into buckets of range 10**

| CLASSIFIER | ACCURACY(%) |
|---|---|
| KNN | 24.51 |
| Naieve Bayes (MNB) | 18.16 |
| Naieve Bayes (GNB) | 74.18 |
| SVM | 47.60 |
| Decision Trees | 27.99 |
| Random Forests | 78.41 |

**Fig 3.b.1.2: Classifier Model's Accuracy for predicting Batsman Dream-11 Scores into buckets of range 10**

The accuracy of predicting a player's Dream-11 score in buckets of 10 is quite acceptable. However, practically speaking, even a difference of 4 points will put us 1000s of rank away from the first optimal fantasy team's rank. Hence, we proposed the method of Regression to predict the Dream-11 scores of the player's performance prediction. Now the job at hand was to find the best of the best regressor models to make predictions for our problem statement.

2) ***Finding Best Prediction Regressor Model***

We had to choose the best regression model for making predictions, and we thought of testing out all sorts of regression models that exists to determine the best. This was a very laborious task. However, luckily, we had PyCaret Library [39] for our rescue.

We sampled 10% of our entire batsman and bowler datasets. We fit it to all the 22 Scikit-Learn's regressor models available in the PyCaret Package's Environment. And we executed the code to compare and evaluate all the models and give us the best Regressor Model with a Maximum acceptable R2 Score. The PyCaret Library did a fascinating job and recommended us the go with the Extra Trees Regressor Model. The recommendation output is the figure below.

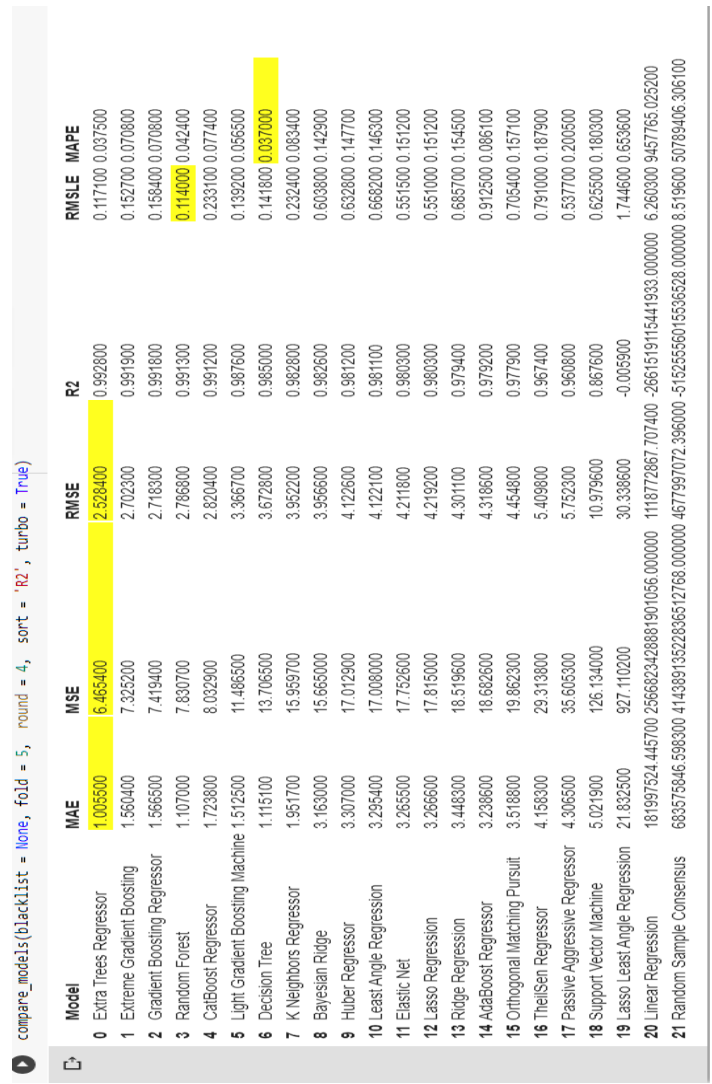

**Fig 3.b.2.1: PyCaret's Recommendation for the Best Learning Prediction Model for our dataset**

We knew that an R2 Score of 0.99 was quite alarming and made us believe that the model was over fitted. But, when we analyzed the case for over-fitting if was surprising that our model did not overfit for the following reasons:
- Learning Rate is higher than Testing rate.
- The correlation between the features chosen and label to be predicted is very high.

The below graphs will support that the model was not over fitted on our dataset.

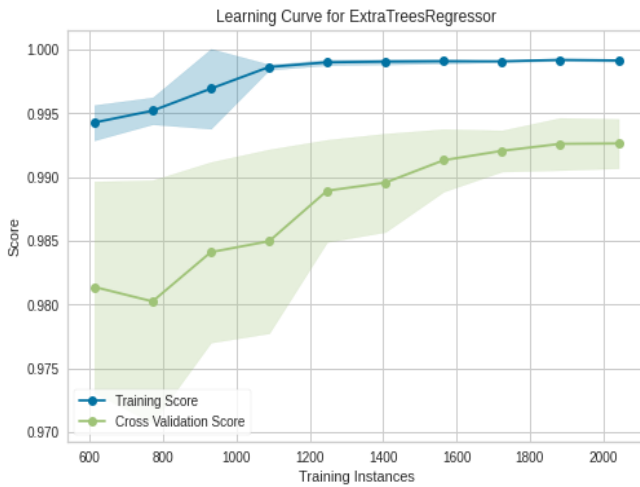

**Fig 3.b.2.2: Learning Curve for ETR Model**

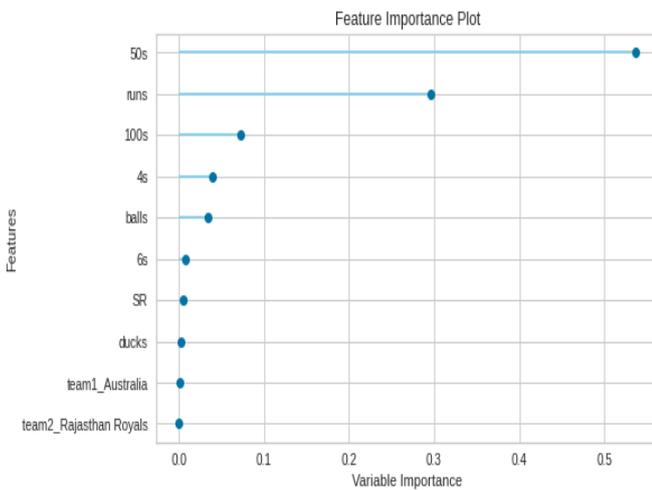

**Fig 3.b.2.3: Batsman Data Feature Importance Plot for ETR Model**

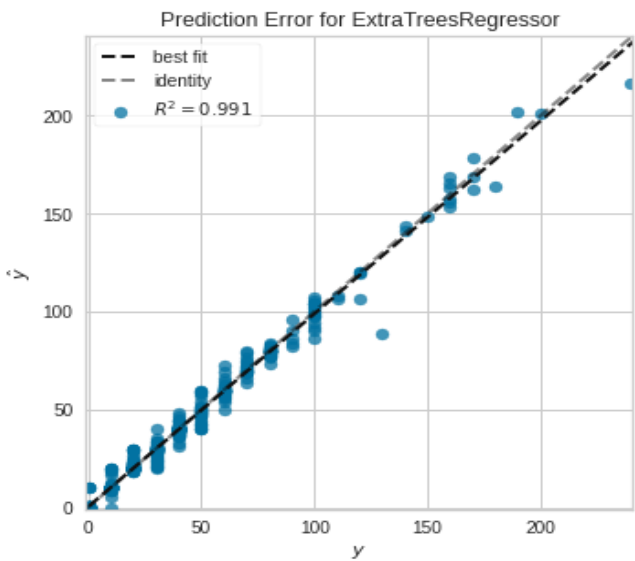

**Fig 3.b.2.4: Prediction Error Plot for ETR Model**

*3) **Building a Real World Usable ML Predictor***

Now after learning that ETR was the best learning prediction model for our dataset, we knew that we had to build our model and store it as a picked file to generate predictions. We sampled 50% of our dataset freshly, and trained our model on this sampled dataset. Then, we fitted the unseen 50% of dataset by our model to make predictions, surprisingly we achieved the almost the same R2 score as predicted by the PyCaret Library. The results are shown below.

```
et.fit(X_train_ohe,y_train)
pred= et.predict(X_test_ohe)
r2 = r2_score(y_test,pred)
print('The r2 using ET is:',format(r2*100))
```

```
/usr/local/lib/python3.6/dist-packages/ipykernel_launcher.py
"""Entry point for launching an IPython kernel.
The r2 using ET is: 99.09200045638103
```

**Fig 3.b.3.1: Prediction R2 Score for ETR Model on Batsman Dream-11 Score Predictions**

```
et.fit(X_train_ohe,y_train)
pred= et.predict(X_test_ohe)
r2 = r2_score(y_test,pred)
print('The r2 using ET is:',format(r2*100))
```

```
/usr/local/lib/python3.6/dist-packages/ipykernel
"""Entry point for launching an IPython kernel
The r2 using ET is: 97.337736076019576
```

**Fig 3.b.3.2: Prediction R2 Score for ETR Model on Bowler Dream-11 Score Predictions**

Our ML Predictor takes the following list of Input values to derive predictions on the Dream-11 Score of the player.

**For Batsman:**
train_feature=['batsman','MF', 'team1', 'team2', 'venue', 'runs', 'balls', '4s', '6s', '50s', '100s', 'ducks', 'SR']

**For Bowler:**
Features=['bowler', 'MF', 'team1', 'team2', 'venue', 'overs', 'runs', 'maidens', 'wicket', 'econrate', '4 wicket', '5 wicket']

However, in real world, it is quite preposterous to expect the user to know all the exact values for these inputs. We can only expect the user to enter the following inputs: [ 'Player', 'MF', 'team1', 'team2', 'venue']. So we need to extract all rows from the dataset that has all these 5 inputs and recursively delete each input feature and extract all rows until all 5 features are deleted and we have a list of rows that can be used as input for our model.

Now, after getting the list of the rows, we must transform the list into a two dimensional matrix that has "n" rows, and "12" columns if the data is extracted from batsman dataset, and "11" columns if the data is extracted from bowler's dataset. Then, we fit the data matrix to the respective prediction models and derive the predictions and give the user the average of all the predictions.

If the player is an all-rounder then, the user will see two prediction values for the player in both batting and bowling format, and his predicted dr11 scores will be summed. Therby, we considered the factors of usability and knowledge of users to derive predictions.

## IV. CONCLUSIONS & FUTURE WORK

- We sampled and used only 10% of the entire dataset to learn the best regression model for our dataset and problem statement using PyCaret.
- The Auto-ML PyCaret library suggested us to go with the **Extra Trees Regressor Model (ETR)** to attain the maximum precise predictions.
- When ETR ML Model was used with the 100% of dataset with a Train-Test Split of 7:3, we got an R2 score of 0.99 for batsman dream-11 scores predictions, and an R2 score of 0.97 for bowler dream-11 scores predictions.
- Hence, with our product we proved that to predict the performance of a player in a prospective game, we must use regressor models and not classification models as suggested by our base paper.

Further, to make our model better, we must construct a pipeline that scrapes data of every new match from the cricsheet website, and perform the transformations, and feature engineering and update the data as rows to our batsman and bowler datasets.